\documentclass[letterpaper]{article} 
\usepackage{aaai24}  
\usepackage{times}  
\usepackage{helvet}  
\usepackage{courier}  
\usepackage[hyphens]{url}  
\usepackage{graphicx} 
\usepackage{natbib}  
\usepackage{caption} 
\usepackage{algorithm}
\usepackage{algorithmic}
\usepackage{amsfonts}
\usepackage{amsmath}
\usepackage{multirow}
\usepackage{xcolor}         

\usepackage{newfloat}
\usepackage{listings}
\DeclareCaptionStyle{ruled}{labelfont=normalfont,labelsep=colon,strut=off} 
\floatstyle{ruled}
\newfloat{listing}{tb}{lst}{}
\floatname{listing}{Listing}
\usepackage{multicol}
\title{Hamiltonian GAN}
\author {
    Christine Allen-Blanchette
}
\affiliations{
  Department of Mechanical and Aerospace Engineering\\
  Center for Statistics and Machine Learning\\
  Princeton University\\
  Princeton, NJ 08450, USA \\
  \texttt{ca15@princeton.edu} \\
}

\usepackage{bibentry}

\begin{document}

\maketitle

\begin{abstract}
A growing body of work leverages the Hamiltonian formalism as an inductive bias for physically plausible neural network based video generation. The structure of the Hamiltonian ensures conservation of a learned quantity (e.g., energy) and imposes a phase-space interpretation on the low-dimensional manifold underlying the input video. While this interpretation has the potential to facilitate the integration of learned representations in downstream tasks, existing methods are limited in their applicability as they require a structural prior for the configuration space at design time. In this work, we present a GAN-based video generation pipeline with a learned configuration space map and Hamiltonian neural network motion model, to learn a representation of the configuration space from data. We train our model with a physics-inspired cyclic-coordinate loss function which encourages a minimal representation of the configuration space and improves interpretability. We demonstrate the efficacy and advantages of our approach on the Hamiltonian Dynamics Suite Toy Physics dataset.

\end{abstract}

\begin{figure*}[]
    \centering
    \includegraphics[width=1\textwidth]{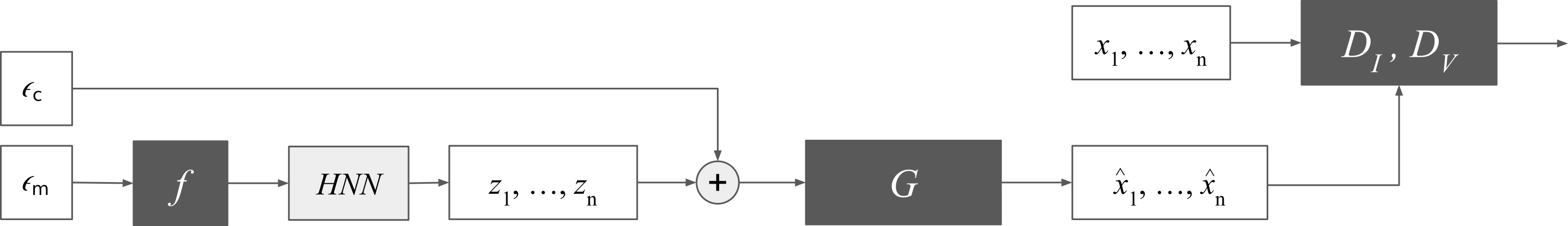}
    \caption{
    HGAN video generation pipeline. Physically plausible videos of conservative systems are generated from motion and content Gaussian random samples $\epsilon_m$ and $\epsilon_c$. The motion sample $\epsilon_m$ is mapped to an initial condition $z_m^0$ of the underlying dynamical system by the configuration space map $f$, then integrated forward in time using Hamilton's equations and the learned Hamiltonian $\mathcal{H}_\theta$ in the HNN module. Each element of the resulting sequence $z_m^t$ is concatenated with the content sample $\epsilon_c$, then passed to the generator network $G_I$ to produce an image $\hat{x}^t$. During training,  individual images are passed to the discriminator $D_I$ and image sequences $\{\hat{x}^j, \dots, \hat{x}^{j+T}\}$ are passed to the discriminator $D_V$.
    }
    \label{fig:my_label}
\end{figure*}

\section{Introduction}
Hamiltonian mechanics can be applied to systems on general manifolds, and is independent of a particular choice of coordinates \cite{holm2009geometric}. This flexibility is attractive in data-driven modelling learning where low-dimensional manifolds are often approximated from high-dimensional data \cite{bengio2013representation}. In this work, we leverage the structure of the Hamiltonian within the generative adversarial network (GAN)~\cite{goodfellow2014generative} framework for physics-guided video generation. The problem of video generation asks for a representation of video frames and frame-to-frame transitions. The Hamiltonian formalism provides a useful framing for this problem: each video frame is an observation of the system state; the set of all states is the phase-space; and state transitions are determined by Hamilton’s equations. With this framing,
the time evolution of the video-generating process is determined by well-understood physical principles which gives an interpretation to the learned representation which can be useful for downstream tasks such as control.

The Hamiltonian formalism has been used as an inductive bias in physics-guided video generation in several recent works~\cite{toth2019hamiltonian,saemundsson2020variational,zhong2020unsupervised,higgins2021symetric}. 
In each of the aforementioned, a variational autoencoder (VAE)~\cite{kingma2013auto} is used as the generative model and a Hamiltonian neural network~\cite{greydanus2019hamiltonian} (HNN) (or symplectic recurrent neural network (SRNN)~\cite{chen2019symplectic}) is used to compute trajectories in the latent space. The use of the HNN model allows for an interpretation of the latent space as a phase-space for the dynamical system underlying the video. While this interpretation has the potential to facilitate the integration of learned representations in downstream tasks such as control, 
a persistent challenge has been in the identification of an appropriate structural prior for the phase-space.

The authors in~\cite{toth2019hamiltonian,saemundsson2020variational,higgins2021symetric}
circumvent this challenge by using a standard Gaussian prior and a phase-space of substantially higher dimension than the dynamical system underlying the video. While this approach is common in the VAE literature, it can lead to solutions with poor reconstructive ability; a limitation discussed in~\cite{davidson2018hyperspherical,dai2019diagnosing} and observed in~\cite{zhong2020unsupervised,botev2021priors}; which can limit the utility of learned representations in downstream tasks. In contrast, the authors in~\cite{zhong2020unsupervised} use prior knowledge of the configuration space to select a distribution with the appropriate structure. While this lends greater interpretability to the latent space, it requires prior knowledge of the structure of the phase-space and the availability of a distribution on that space that admits the reparameterization trick~\cite{kingma2013auto}. 

In our work, we use a generative adversarial network (GAN)~\cite{goodfellow2014generative} based video generation approach to circumvent the challenge of identifying an appropriate distribution in advance by implicitly learning the structure of the phase-space from data. We take inspiration from other GAN-based video generation pipelines~\cite{tulyakov2018mocogan,yoon2019time,gordon2021latent,skorokhodov2021stylegan} where motion and content vectors are initialized with Gaussian random noise, propagated forward in time using a recurrent neural network, and mapped into the image space by a generator network. To allow for the interpretation of the motion vectors as elements of phase-space for the underlying dynamical system, we incorporate three key elements: (1) a configuration-space map that learns to transform Gaussian random vectors to an intermediate space which we interpret as the configuration space of the system; (2) a HNN module in place of the standard discrete recurrent unit to enforce continuous and conservative latent dynamics, and (3) a physics-inspired cyclic-coordinate loss function that encourages a sparsity in the configuration space representation allowing for the discovery of phase-space structure.

We empirically evaluate our model on two versions of the Toy Physics dataset~\cite{toth2019hamiltonian}, one with constant physical parameters and colors and another where the physical parameters and colors vary.
Our analysis highlights the importance of the HNN module for temporal coherence of generated sequences, and of the configuration-space map and cyclic-coordinate loss for a compact and interpretable representation of the motion subspace.
\section{Related work} 
In the physics-guided learning literature, 
a growing number of works investigate the integration of Hamiltonian and Lagrangian formalisms as an inductive bias for dynamical systems forecasting. The authors in~\cite{greydanus2019hamiltonian} learn to predict the change in position and momentum for conservative mechanical systems using a system specific Hamiltonian parameterized by a neural network.
The authors in~\cite{chen2019symplectic} find that using the Hamiltonian neural network as a recurrent module improves long term prediction accuracy without the need for gradient information. This observation is leveraged in~\cite{zhong2019symplectic,lutter2019deep,roehrl2020modeling} for control, ~\cite{allen2020lagnetvip,zhong2020unsupervised} for video prediction and ~\cite{toth2019hamiltonian,saemundsson2020variational} for variational autoencoding (VAE)~\cite{kingma2013auto} based video generation. To the best of our knowledge, we are the first to use the Hamiltonian or Lagrangian formalism as an inductive bias in generative adversarial network (GAN)~\cite{goodfellow2014generative} based video generation. 
In contrast to VAE-based video generation methods which impose an explicit prior,
GAN-based video generation methods implicitly learn the data distribution during training. The authors in~\cite{vondrick2016generating} learn to map Gaussian random noise vectors to image sequences using a generator network with 3D deconvolutional layers. The authors in~\cite{tulyakov2018mocogan,yoon2019time} find they are able to substantially reduce complexity and improve performance by decomposing the Gaussian random noise vector into motion and content vectors, propagating them forward in time using a recurrent neural network, then mapping each resulting vector to an image using a generator network with 2D deconvolutional layers. This high-level structure is used in many recent GAN-based video generation methods. Some of these methods introduce alternatives to the standard convolutional generator network -- the authors in~\cite{yu2022generating} use alternative architecture types, the authors in~\cite{tian2021good} use large pre-trained image generator networks, and the authors in~\cite{wang2020g3an} use multiple generator networks. Others have worked to increase the expressivity of the latent code -- the authors in~\cite{wang2020g3an} use a content-motion integration scheme to improve latent space disentanglement, and the authors in ~\cite{skorokhodov2021stylegan} learn a map from the standard Gaussian distribution to an intermediate distribution which may better represent the data distribution. Still others have introduced alternative motion models -- the authors in~\cite{tian2021good} use an RNN-based motion model to learn dynamical updates rather than subsequent states, and the authors in~\cite{gordon2021latent} replace the discrete recurrent unit with the continuous NeuralODE~\cite{chen2018neural}.  

The work in~\cite{toth2019hamiltonian} and \cite{gordon2021latent} are most similar to ours. \cite{toth2019hamiltonian} use the Hamiltonian formalism as an inductive bias for VAE-based video generation. Their framework generates physically plausible video sequences at different time scales and time directions. We also use a Hamiltonian neural network for video generation and thereby inherit these capabilities; by using the GAN framework, however, we circumvent the challenge of selecting an appropriate prior distribution in advance of training. 
\cite{gordon2021latent} use the Neural ODE~\cite{chen2018neural} recurrent unit to learn continuous dynamics in a GAN-based video generation pipeline. They demonstrate performance commensurate with other GAN based video generation approaches while benefiting from a continuous motion model. Our motion model is distinct from~\cite{gordon2021latent}, however, since we enforce continuous and conservative dynamics on a transformed latent space which we interpret as the underlying configuration space. Moreover, we impose a novel cyclic-coordinate loss function to encourage sparsity in the configuration space representation to facilitate discovery of phase-space structure. 

\section{Background}
In this section we present Hamilton's equations and describe how they can be used to identify cyclic coordinates. We also present Hamiltonian neural networks~\cite{greydanus2019hamiltonian}, generative adversarial neural networks~\cite{goodfellow2014generative}, and the GAN-based video generation approach MoCoGAN~\cite{tulyakov2018mocogan} as themes from each of these appear in the proposed approach. 

\subsection{Hamilton's equations and cyclic coordinates}
\label{sec:cyclic_coords}
The Hamiltonian formalism is a reformulation of Newton's second law, $F=ma$, in terms of the system energy rather than its forces~\cite{goldstein2002classical}. This formulation is preferable in settings such as ours where the identification and quantification of system forces may be challenging. The Hamiltonian of a system, $\mathcal{H}:T^*\mathcal{Q}\mapsto \mathbb{R}$, is a map from generalized position and momentum coordinates $(q,p)\in T^*\mathcal{Q}$, to the total energy of the system, where $T^*\mathcal{Q}$ denotes the co-tangent space of the system configuration space $\mathcal{Q}$. The behavior of a Hamiltonian system is governed by Hamilton's equations: 
\begin{equation}
    \label{eqn:hamiltons_eqns}
    \dot{p} = -\frac{\partial E}{\partial q}, \; \dot{q} = \frac{\partial E}{\partial p}, \quad E=\mathcal{H}(q,p).
\end{equation}
Given the system Hamiltonian and its initial conditions, we can forecast future system states by integrating Hamilton's equations forward in time. 

We can also use Hamilton's equations to identify cyclic coordinates. The generalized position coordinate $q_i$ is called cyclic or ignorable if it does not contribute to the total energy of the system:
\begin{equation}
\label{eq:cyclic-coord}
-\frac{\partial E}{\partial q_i} = 0 = \dot{p_i}.
\end{equation}
From equation \ref{eq:cyclic-coord}, we see that for a cyclic coordinate $q_i$, the corresponding conjugate momentum $p_i$ is conserved~\cite{goldstein2002classical}.

Previous work using Hamilton's equations for dynamical systems forecasting either assume prior knowledge of the dimension of the configuration space for improved interpretability~\cite{allen2020lagnetvip,zhong2020unsupervised}, or select the dimension to be arbitrarily large for model flexibility~\cite{toth2019hamiltonian,saemundsson2020variational}. In our approach we achieve interpretability with an arbitrarily large latent dimension by incorporating our novel cyclic-coordinate loss function. Our loss function, detailed in the section: Hamiltonian GAN, encourages the identification of cyclic coordinates by regularizing the change in momentum.

\subsection{Hamiltonian Neural Networks}
Hamiltonian neural networks (HNNs)~\cite{greydanus2019hamiltonian,chen2019symplectic} model the underlying dynamics of sequential data using the Hamiltonian formalism. The data are assumed to be sampled from a continuous and conservative dynamical system, and the system Hamiltonian is assumed to be unknown. 

HNNs are one of a larger class of models termed NeuralODEs~\cite{chen2018neural}, that use neural networks to parameterize the ordinary differential equations (ODEs) governing the dynamics underlying sequential data. 
NeuralODEs learn a neural network $f_\theta$, with parameters $\theta$, to predict a next state, $x(t_1)$, from a current state $x(t_0)$ by the following:
\begin{equation*}
x(t_1) = x(t_0) + \int_{t_0}^{t_1} f_\theta(x(t_0), t)\, dt, \quad \frac{d x(t)}{d t} = f_\theta(x(t), t).
\end{equation*}
In HNNs, the system Hamiltonian is parameterized using a neural network and next state predictions are formed by integrating Hamilton's equations (equation~\ref{eqn:hamiltons_eqns}). 

HNNs have been used for dynamical systems forecasting when position-momentum values are assumed to be known~\cite{greydanus2019hamiltonian,chen2019symplectic} and in VAE-based video generation pipelines where ground truth position-momentum values are assumed to be unknown~\cite{toth2019hamiltonian,saemundsson2020variational}. As far as we know, we are the first to use HNNs for GAN-based video generation. 

\subsection{Generative Adversarial Networks (GANs)}
In the generative adversarial network (GAN)~\cite{goodfellow2014generative} framework two networks -- the generator network $G$, and discriminator network $D$ -- are learned with opposing objectives. The objective of the generator is to learn to map samples from a known distribution $p_\text{z}(z)$, to a distribution that resembles the training data distribution $p_{\text{data}}(x)$. The objective of the discriminator is to discern whether or not a given sample is from the training set and indicate this with a 1 or 0 respectively. The formulation of these contrasting objectives proposed in \cite{goodfellow2014generative}, is the following:
\begin{align*}
    \min_G \max_D \mathcal{L}(D, G) &=  \mathbb{E}_{z\sim p_\text{z}(z)}\log[ 1 - D(G(z)) ]\\&+ \mathbb{E}_{x\sim p_{\text{data}}(x)}[\log D(x)].
\end{align*}
The objective is maximized when the discriminator correctly labels synthetic samples, $G(z)$, with zero and samples from the training dataset, $x$, with one. The objective is minimized when the generator produces synthetic samples the discriminator incorrectly labels with one.

\subsection{Motion Content GAN (MoCoGAN)}
\label{sec:mocogan}
MoCoGAN~\cite{tulyakov2018mocogan} applies the GAN framework to video generation. The processing pipeline begins with sampling a Gaussian random noise vector $z$ and decomposing it into a motion vector $z_m^0$ and content vector $z_c$. The motion vector is passed to a recurrent neural network $R$, which produces a the motion vector sequence $\{z_m^j\}_{j=0}^N$, where $N$ is the sequence length. Content and motion vectors are concatenated to form the vector sequence $\{z_j\}_{j=0}^N$, where $z_j = (z_c, z_m^j)$, and each vector $z_j$ is mapped to an image $\tilde{x}_j$ by a 2D deconvolutional generator network $G_I$. The resulting image sequence $\{\tilde{x}_j\}_{j=0}^N$ is interpreted as a synthetic video, where each image $\tilde{x}_j$ is interpreted as a video frame. 

During training, an image discriminator network $D_I$ predicts whether or not an input image is from the training set, and a video discriminator network $D_V$ predicts whether or not a video is from the training set. The generator and discriminator networks are trained under the objective function:
\begin{align*}
\min_{G_I, R} &\max_{D_I, D_V} \mathcal{L}(D_I, D_V, G_I, R) = E_{\tilde{v}}[\log(1-D_I(S_1(\tilde{v})))] \\&+ E_v[\log D_I(S_1(v))] + E_{\tilde{v}}[\log(1-D_V(S_T(\tilde{v})))] \\&+ E_v[\log D_V(S_T(v))]
\end{align*}
where $v$ denotes a video from the training dataset, $\tilde{v}$ denotes a synthetic video, and $S_i$ denotes the random sampling of $i$ consecutive video frames.

MoCoGAN improved upon previous approaches by modeling motion and content separately, and generating trajectories in the latent space. To allow for the interpretation of the motion vectors as elements of phase-space for the underlying dynamical system, we incorporate a configuration space map $f$ that learns to transform Gaussian random vectors to an intermediate space which we interpret as the configuration space of the system, and an HNN module in place of the discrete recurrent unit to enforce continuous and conservative latent dynamics.

\section{Hamiltonian GAN}
In this section we detail our model, a three-stage GAN-based video generation pipeline in which a configuration-space map and cyclic-coordinate loss function encourage identification of a minimal configuration-space; and a HNN motion model supports generation of physically plausible video.
\subsection{Video generation}
Our inference framework consists of a configuration-space map $f:\mathbb{R}^n\mapsto \mathbb{R}^k$, an HNN~\cite{chen2019symplectic} motion model with a learned Hamiltonian $\mathcal{H}:T^*\mathcal{Q}\mapsto \mathbb{R}$, and an image generator $G_I:\mathcal{Z}\mapsto\mathcal{I}$. Similarly to MoCoGAN~\cite{tulyakov2018mocogan} we begin by sampling a Gaussian random noise vector $z$ and decomposing it into a motion vector $z_m$ and content vector $z_c$. In contrast to MoCoGAN, however, we assume the structure of the motion subspace can be learned by our piecewise continuous configuration map $f$, and that the underlying dynamical system is continuous and conservative.

Given the Gaussian sample $z_m$, the configuration-space map $f$, a two layer MLP, outputs a vector $y_0$ which is interpreted as an initial condition on the phase-space $T^*\mathcal{Q}$. This initial condition is passed to the HNN motion model which computes
future system states $y_j=(q_j, p_j)$ using Hamilton's equations (equation~\ref{eqn:hamiltons_eqns}) and the learned Hamiltonian $\mathcal{H}$. Concretely, the sequence of motion vectors $\{y_j\}_{j=0}^{N-1}$, where $N$ is the sequence length, is given by:
\begin{equation}
\label{eq:integration}
    y_0 = f(z_m); \quad y_j = y_{j-1} + \int_{t_{j-1}}^{t_j} \dot{y}_{j-1}\, dt, \; j>0
\end{equation}
where $\dot{y}_j = (\dot{q}_j, \dot{p}_j)$
is given by Hamilton's equations (equation~\ref{eqn:hamiltons_eqns}). In our implementation, we compute the integral~\ref{eq:integration} using leapfrog integration with $dt=0.05$. While a simpler integration scheme can be used, leapfrog integration preserves simplicity and has been shown to outperform other integration techniques~\cite{hairer2006geometric}. 

Given the content vector $z_c$ and motion sequence $\{y_j\}_{j=0}^{N-1}$, we form the latent sequence $\{w_j\}_{j=0}^{N-1}$ where $w_j=(z_c,\, y_j)$. Given this latent sequence, we generate a synthetic video $\tilde{v}=\{\tilde{x}_j\}_{j=0}^{N-1}$ where images $\tilde{x}_j = G_I(w_j)$, are determined by the generator network.

\subsection{Cyclic-coordinate loss}
\label{sec:cyclic-coordinate_loss}
Our configuration-space map $f$ defines an intermediate space between the sampling distribution, and the motion model determined by the learned Hamiltonian $\mathcal{H}$. We interpret this intermediate space as a configuration space for the dynamical system underlying the video data. Since its dimension is selected arbitrarily at design time, without constraint, the learned representation may be dispersed with little identifiable structure. To encourage a minimal representation and improve interpretability, we introduce a cyclic-coordinate loss function inspired by the observation that the momentum coordinate $p_i$, corresponding to a cyclic-coordinate $q_i$ is conserved (see section: Background). We define our loss:
\begin{equation}
\label{eq:cyclic-coord-loss}
    \mathcal{L}_{cyc} = \frac{1}{N}\sum_i \lambda\; |\dot{p}_i|,
\end{equation}
where $N$ is the batch size and $i$ ranges over the latent dimension. 
We choose a regularization penalty of $\lambda=0.01$ for all experiments. Our full model is trained end-to-end using the MoCoGAN loss (see section: Background) amended with our cyclic-coordinate loss (equation~\ref{eq:cyclic-coord-loss}). 

\begin{table*}
\caption{FVD comparison on the CCC Toy Physics dataset. 
The consistent advantage of our model over MoCoGAN suggests that the HNN recurrent module is able to generate latent trajectories on the motion manifold more reliably than the GRU module. The advantage of our model over the HGN model suggests that the implicitly learned configuration space structure is a better representation of the actual configuration space than the Gaussian prior used in HGN. The best score is shown in bold, the second best is shown in blue.}
\label{tab:fvd_ccc}
\centering
\begin{tabular}{ | l | c | c | c | c | c |}
\hline
  & \multirow{2}{3em}{Mass spring} & \multirow{2}{4em}{Pendulum} & \multirow{2}{4em}{Double Pendulum} & \multirow{2}{3em}{Two-body} & \multirow{2}{3em}{Three-body} \\ 
  & & & & & \\
  \hline
 HGN  & 385.08  & 688.12 & \color{blue}{331.94} & \color{blue}{830.91} & \textbf{451.40} \\
 \hline 
 MoCoGAN & \textbf{31.37}  & \color{blue}{398.12} & 979.80 & 1028.13 & 2527.78\\ 
  \hline
 HGAN (Ours) & \color{blue}{45.68} & \textbf{91.64} & \textbf{73.21} & \textbf{105.85} & \color{blue}{1981.10}\\  
  \hline 
\end{tabular}
\end{table*}
\section{Empirical analysis}
\label{sec:results}
In this section we compare 
videos generated using 
the following models:
\begin{enumerate}
    \item HGN~\cite{toth2019hamiltonian}: A VAE-based video generation approach in which the latent vector is interpreted as an element of the phase-space and propagated forward in time with an HNN~\cite{greydanus2019hamiltonian} cell. We use the pytorch implementation of HGN introduced in~\cite{rodas2021re}.
    \item MoCoGAN~\cite{tulyakov2018mocogan}: A GAN-based video generation approach in which content and motion vectors are initialized with Gaussian random noise. A motion sequence is generated using a discrete time RNN cell, and a latent sequence is constructed by concatenating the content vector to each motion vector. Each latent vector is decoded by the generator network to produce a sequence of images. We use the pytorch implementation of MoCoGAN introduced in~\cite{mocogan-pytorch}.
    \item HNN-GAN: This model differs from MoCoGAN in that the discrete time RNN cell is replaced with a continuous time HNN cell. This model can be seen as a variant of HGAN (our model), in which the configuration space map is ablated.
    \item HGAN (Ours): This model differs from HNN-GAN in that a configuration space map transforms the Gaussian random noise initialized motion vector to an intermediate space. The resulting vector is interpreted as the initial condition for the HNN cell and propagated forward in time to give a motion sequence. Moreover, a cyclic-coordinate loss function is used during training to encourage a compact representation of the configuration-space.
\end{enumerate} 

All models are trained on the Hamiltonian Dynamics Suite Toy Physics dataset~\cite{botev2021priors}. HGN and MoCoGAN are trained using the hyperparameters provided in the original paper. The hyperparameters for HNN-GAN and HGAN (Ours) are optimized as described in the Appendix.

\subsection{Datasets}
We compare model performance on two variations of the Hamiltonian Dynamics Suite Toy Physics dataset~\cite{botev2021priors}. The dataset consists of a collection of video renderings of simulated physical systems. It includes rendered simulations of the mass spring, double pendulum, pendulum, two-body and three-body systems. Simulations are performed with and without friction; and videos are rendered with constant or varied physical quantities (e.g., the mass of the pendulum bob, the length of the pendulum arm, and location of the pivot in the pendulum dataset) across trajectories, and constant grayscale, constant color, or varied color across trajectories. Each simulation is sampled at $dt=0.05$, and 50k trajectories of 512 samples are generated for every system. Further details about the dataset can be found in~\cite{botev2021priors}. 

Following HGN we evaluate all models on the version of this dataset generated without friction, with constant physical quantities across trajectories, and with constant color. We perform additional experiments using the version of this dataset generated without friction, with constant physical quantities, and with varied color. We label these datasets CCC and CCV in our results. The CCV dataset does not include rendered simulations of the three-body system.
\begin{table}[b]
\caption{FVD comparison on the CCV Toy Physics dataset. The consistent advantage of our model over MoCoGAN suggests that the HNN recurrent module is able to generate latent trajectories on the motion manifold more reliably than the GRU module. Our model shows comparable performance to the HNN-GAN model. The best score is shown in bold, the second best is shown in blue.}
\label{tab:fvd_cvv}
\centering
\begin{tabular}{ | l | c | c | c | c |}
\hline
  & \multirow{2}{3em}{Mass spring} & \multirow{2}{4em}{Pendulum} & \multirow{2}{4em}{Double Pendulum} & \multirow{2}{3em}{Two-body}  \\ 
  & & & & \\
 \hline 
 MoCoGAN & \color{blue}{736.01} & 1030.21 & \textbf{803.02} &  716.86\\ 
  \hline
 HNN-GAN & 787.42 & \textbf{618.33} & \color{blue}{625.75} & \textbf{417.92}\\ 
  \hline
 HGAN (Ours) & \textbf{436.80} & \color{blue}{816.04} & 1124.56 & \color{blue}{444.02} \\  
  \hline 
\end{tabular}
\end{table}
\begin{figure*}
    \centering
    \includegraphics[width=\textwidth]{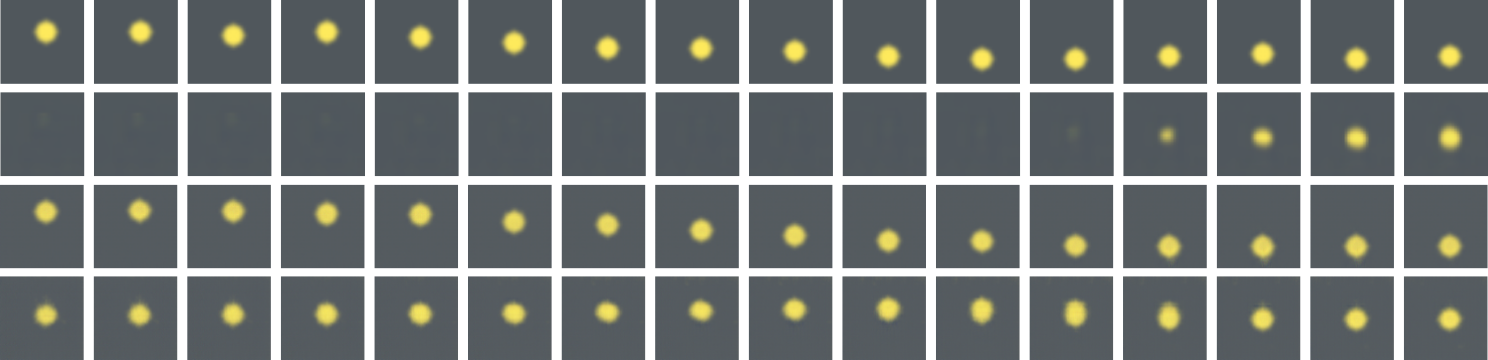}
    \caption{Unconditional mass spring video generation (CCC Toy physics dataset). We show random samples from the training set, HGN, MoCoGAN, and our model (top-bottom).}
    \label{fig:synthetic_mass_spring_ccc}
\end{figure*}
\begin{figure*}
    \centering
    \includegraphics[width=\textwidth]{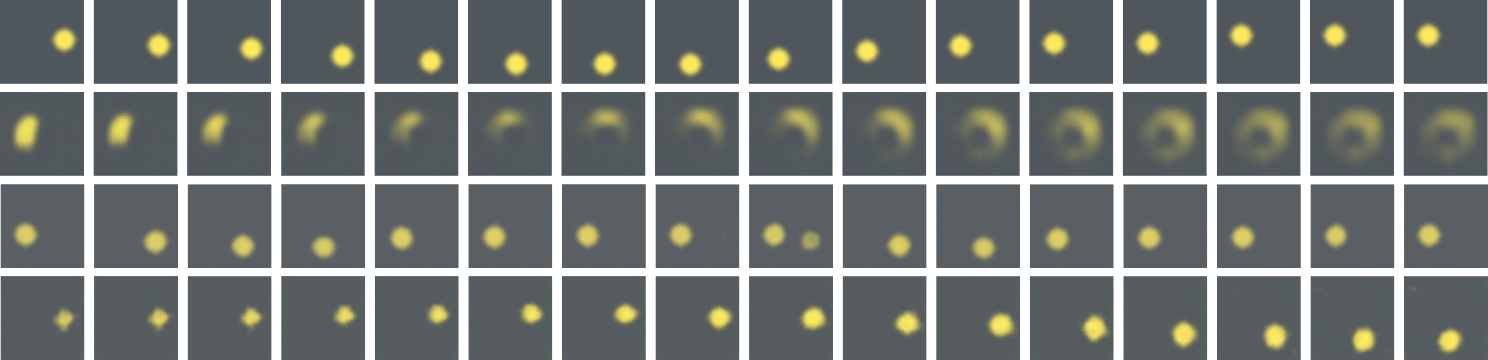}
    \caption{Unconditional pendulum video generation (CCC Toy physics dataset). We show random samples from the training set, HGN, MoCoGAN, and our model (top-bottom).}
    \label{fig:synthetic_pendulum_ccc}
\end{figure*}
\begin{figure*}
    \centering
    \includegraphics[width=\textwidth]{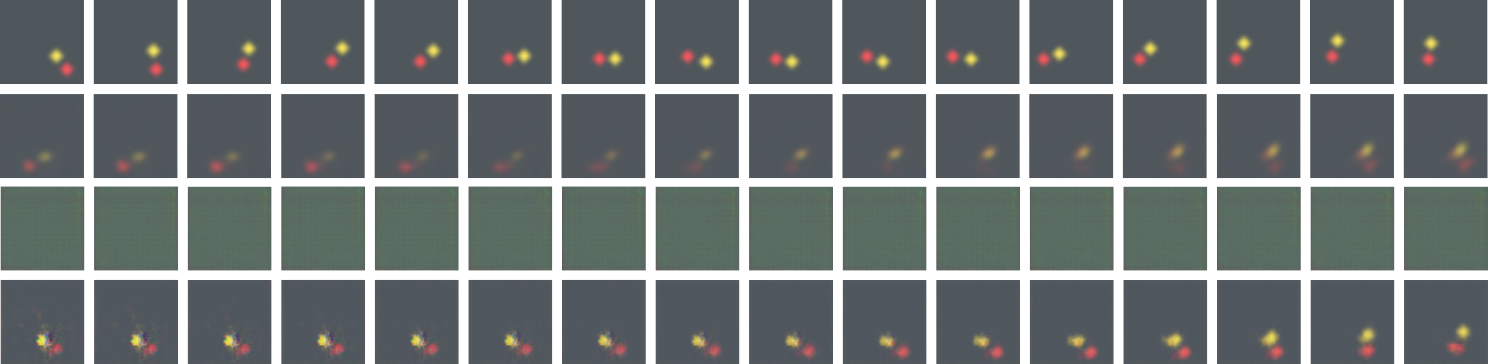}
    \caption{Unconditional double pendulum video generation (CCC Toy physics dataset). We show random samples from the training set, HGN, MoCoGAN, and our model (top-bottom).}
    \label{fig:synthetic_double_pendulum_ccc}
\end{figure*}
\begin{figure*}
    \centering
    \includegraphics[width=\textwidth]{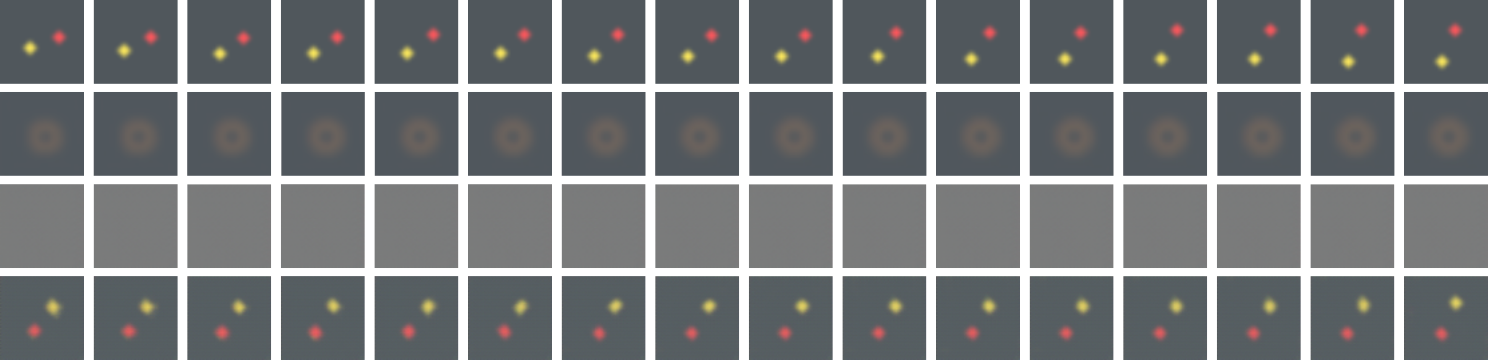}
    \caption{Unconditional two body video generation (CCC Toy physics dataset). We show random samples from the training set, HGN, MoCoGAN, and our model (top-bottom).}
    \label{fig:synthetic_two_body_ccc}
\end{figure*}
\begin{figure*}
    \centering
    \includegraphics[width=\textwidth]{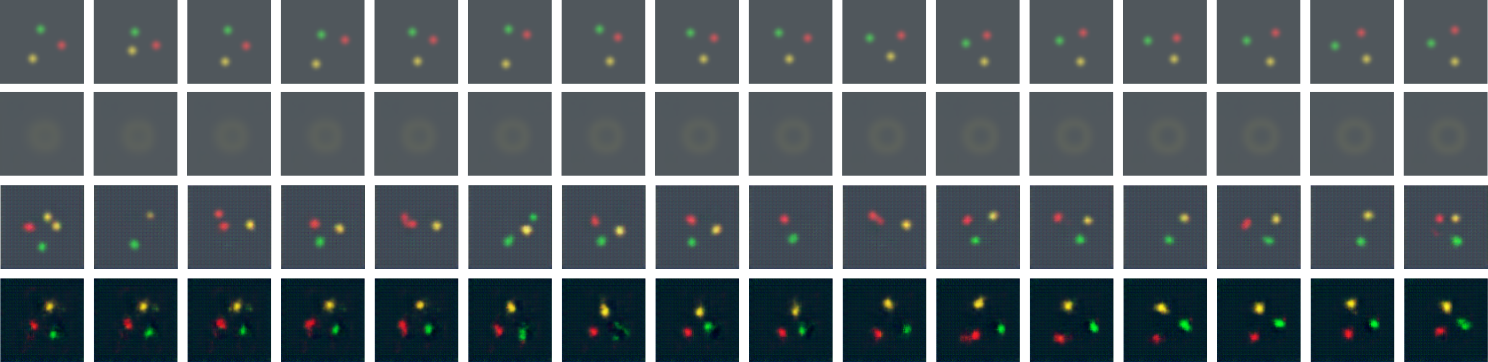}
    \caption{Unconditional three-body video generation (CCC Toy physics dataset). We show random samples from the training set, HGN, MoCoGAN, and our model (top-bottom).}
    \label{fig:synthetic_three_body_ccc}
\end{figure*}
%
\subsection{Perceptual quality}
We compare the performance of our model against the baseline models both quantitatively and qualitatively. We quantitatively compare the perceptual quality of generated videos using the Fr\'echet Video Distance (FVD)~\cite{unterthiner2019fvd}.  
To avoid discrepancies in the evaluation due to subsampling and data processing procedures, we use FVD evaluation pipeline proposed in~\cite{skorokhodov2021stylegan}, and compute the FVD score for each model using 2048 videos of 16 frames. 
We compare the perceptual quality of videos qualitatively by randomly generating 5 video sequences for each model and selecting the best for comparison. Our random seed is set to 0 for all models.

We compare the performance of our model against the HGN, and MoCoGAN baseline models on the CCC dataset. We randomly generate synthetic video sequences using each model and compare them in Figures ~\ref{fig:synthetic_mass_spring_ccc}-\ref{fig:synthetic_three_body_ccc}; we report the FVD scores for each model in Table~\ref{tab:fvd_ccc}. Our model out performs the baseline models in three out of five cases (i.e., Pendulum, Double Pendulum, Two-body), and demonstrates comparable performance to the leading model (MoCoGAN) in one of the remaining cases (i.e., Mass Spring). There is a significant difference in the FVD score of the leading model (HGN), and our model on the Three-body case. 
As evidenced by Figure~\ref{fig:synthetic_three_body_ccc}, our model produces qualitatively better videos than HGN. We attribute the discrepancy in FVD score to the difference in the background color of videos generated with our model and HGN. While videos generated with our model have a foreground more similar to that of the real data, those generated with HGN have a more similar background. 

We compare the performance of our model against the MoCoGAN, and HNN-GAN baseline models on the CCV dataset. Note that we do not compare against HGN since it is designed to model a single system, and color is considered a system parameter. We randomly generate synthetic video sequences using each model and compare them in section CCV of the supplementary material; we report the FVD scores for each model in Table~\ref{tab:fvd_cvv}. Our model out performs baseline models in the Mass Spring case; and performs comparably to the baseline models in the Pendulum and Two-body cases. There is a significant difference in the FVD score of the leading model (HNN-GAN) and our model on the Double Pendulum case. In this case, as evidenced by the figure in section CCV of the supplementary material, our generator network failed during training.

%
%
\begin{figure}[b!]
\centering
    \includegraphics[width=0.4\textwidth]{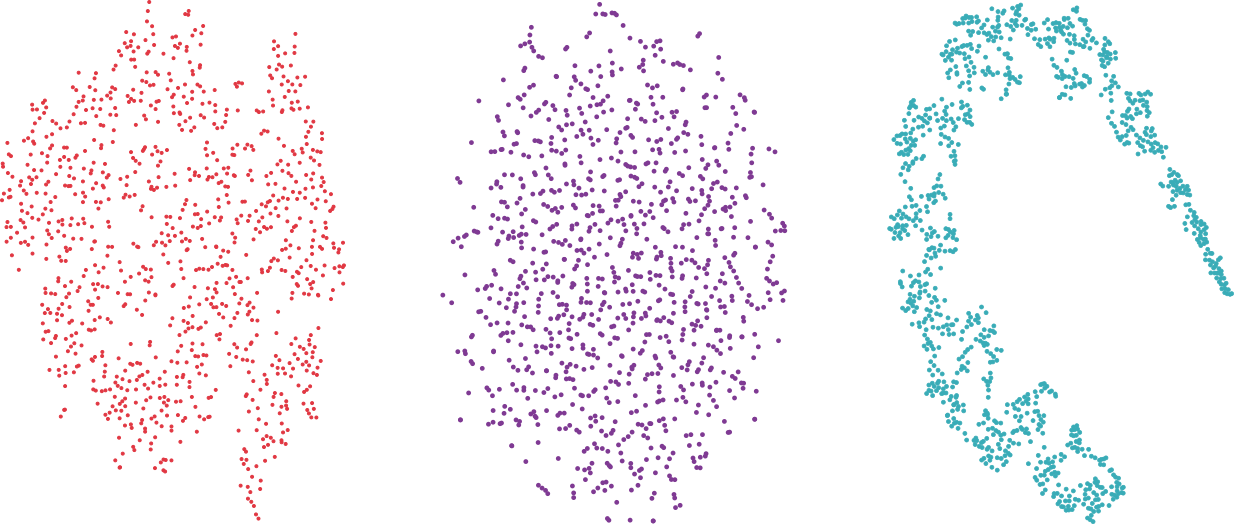}
    \caption{t-SNE projection of the motion manifold. We propagate the Gaussian random motion vector through the synthesis pipelines of MoCoGAN, HNN-GAN, and our model (left-right), and visualize the one step output of the motion module.}
    \label{fig:configuration_space}
\end{figure}
\subsection{Configuration space structure}
We investigate the impact of the configuration space map, and cyclic-coordinate loss on the learned latent structure. We sample 1024 Gaussian random motion vectors and propagate them through the synthesis pipelines of MoCoGAN, HNN-GAN, and our model trained on the Pendulum CCC Toy Physics dataset. We visualize the motion manifold using t-SNE~\cite{van2008visualizing} and PACMAP~\cite{wang2020understanding} projections. Since t-SNE projections are sensitive to the choice of perplexity~\cite{wattenberg2016how} we compute t-SNE projections with perplexity (2, 5, 30, 50, 100) and present the most stable projection in Figure~\ref{fig:configuration_space}. Visualizations of the MoCoGAN and HNN-GAN motion manifolds show no discernible structure, while the motion manifold for our model suggests the system is one dimensional.
\section{Conclusion}
In this work, we introduce a video generation model for physical systems that does not require an explicit prior on the structure of the configuration space. We learn a representation of the underlying configuration space, and leverage the Hamiltonian formalism to learn a continuous and conservative motion model. While the latent dimension of our model exceeds the dimension of the systems considered, we are able to learn compact representations of the configuration space because of our physics-inspired cyclic-coordinate loss function. Our model outperforms baselines in many cases, permits modeling of multiple systems simultaneously, and affords practitioners a mechanism for investigating the learned configuration space structure.

\paragraph{Limitations.}
The proposed Hamiltonian GAN incorporates a Hamiltonian neural network which is unable to capture the dynamics of non-conservative systems. Future work might explore leveraging the Port Hamiltonian formulation as in~\cite{zhong2020dissipative} to model systems with friction. 
The proposed method, is also GAN-based. While this has its benefits, i.e., the data distribution is learned implicitly, there are also detriments, i.e., GAN training can be unstable and slow, and the models learned can suffer from mode collapse. We observed training instability in our model while learning to generate videos of the chaotic double pendulum system (see Figures \ref{fig:synthetic_double_pendulum_ccc}, and \ref{fig:synthetic_double_pendulum_cvv}).
To evaluate the perceptual quality of videos generated with our approach and the baseline approaches, we use FVD~\cite{unterthiner2019fvd}. While the FVD metric aims to capture image quality, temporal coherence and sample diversity, it does exhibit bias~\cite{skorokhodov2021stylegan}. We choose to use this metric, however, as it is standard. We discuss discrepancies between FVD scores and human perceived video quality in Section\ref{sec:results}. 

\bibliography{aaai24}

\begin{thebibliography}{35}
\providecommand{\natexlab}[1]{#1}

\bibitem[{Allen-Blanchette et~al.(2020)Allen-Blanchette, Veer, Majumdar, and
  Leonard}]{allen2020lagnetvip}
Allen-Blanchette, C.; Veer, S.; Majumdar, A.; and Leonard, N.~E. 2020.
\newblock Lagnetvip: A lagrangian neural network for video prediction.
\newblock \emph{arXiv preprint arXiv:2010.12932}.

\bibitem[{Bengio, Courville, and Vincent(2013)}]{bengio2013representation}
Bengio, Y.; Courville, A.; and Vincent, P. 2013.
\newblock Representation learning: A review and new perspectives.
\newblock \emph{IEEE transactions on pattern analysis and machine
  intelligence}, 35(8): 1798--1828.

\bibitem[{Botev et~al.(2021)Botev, Jaegle, Wirnsberger, Hennes, and
  Higgins}]{botev2021priors}
Botev, A.; Jaegle, A.; Wirnsberger, P.; Hennes, D.; and Higgins, I. 2021.
\newblock Which priors matter? Benchmarking models for learning latent
  dynamics.

\bibitem[{Chen et~al.(2018)Chen, Rubanova, Bettencourt, and
  Duvenaud}]{chen2018neural}
Chen, R.~T.; Rubanova, Y.; Bettencourt, J.; and Duvenaud, D.~K. 2018.
\newblock Neural ordinary differential equations.
\newblock \emph{Advances in neural information processing systems}, 31.

\bibitem[{Chen et~al.(2019)Chen, Zhang, Arjovsky, and
  Bottou}]{chen2019symplectic}
Chen, Z.; Zhang, J.; Arjovsky, M.; and Bottou, L. 2019.
\newblock Symplectic recurrent neural networks.
\newblock \emph{arXiv preprint arXiv:1909.13334}.

\bibitem[{Dai and Wipf(2019)}]{dai2019diagnosing}
Dai, B.; and Wipf, D. 2019.
\newblock Diagnosing and enhancing VAE models.
\newblock \emph{arXiv preprint arXiv:1903.05789}.

\bibitem[{Davidson et~al.(2018)Davidson, Falorsi, De~Cao, Kipf, and
  Tomczak}]{davidson2018hyperspherical}
Davidson, T.~R.; Falorsi, L.; De~Cao, N.; Kipf, T.; and Tomczak, J.~M. 2018.
\newblock Hyperspherical variational auto-encoders.
\newblock \emph{arXiv preprint arXiv:1804.00891}.

\bibitem[{Goldstein, Poole, and Safko(2002)}]{goldstein2002classical}
Goldstein, H.; Poole, C.; and Safko, J. 2002.
\newblock Classical mechanics.

\bibitem[{Goodfellow et~al.(2014)Goodfellow, Pouget-Abadie, Mirza, Xu,
  Warde-Farley, Ozair, Courville, and Bengio}]{goodfellow2014generative}
Goodfellow, I.; Pouget-Abadie, J.; Mirza, M.; Xu, B.; Warde-Farley, D.; Ozair,
  S.; Courville, A.; and Bengio, Y. 2014.
\newblock Generative adversarial nets.
\newblock \emph{Advances in neural information processing systems}, 27.

\bibitem[{Gordon and Parde(2021)}]{gordon2021latent}
Gordon, C.; and Parde, N. 2021.
\newblock Latent Neural Differential Equations for Video Generation.
\newblock In \emph{NeurIPS 2020 Workshop on Pre-registration in Machine
  Learning}, 73--86. PMLR.

\bibitem[{Greydanus, Dzamba, and Yosinski(2019)}]{greydanus2019hamiltonian}
Greydanus, S.; Dzamba, M.; and Yosinski, J. 2019.
\newblock Hamiltonian neural networks.
\newblock \emph{Advances in Neural Information Processing Systems}, 32.

\bibitem[{Hairer et~al.(2006)Hairer, Hochbruck, Iserles, and
  Lubich}]{hairer2006geometric}
Hairer, E.; Hochbruck, M.; Iserles, A.; and Lubich, C. 2006.
\newblock Geometric numerical integration.
\newblock \emph{Oberwolfach Reports}, 3(1): 805--882.

\bibitem[{Higgins et~al.(2021)Higgins, Wirnsberger, Jaegle, and
  Botev}]{higgins2021symetric}
Higgins, I.; Wirnsberger, P.; Jaegle, A.; and Botev, A. 2021.
\newblock SyMetric: Measuring the Quality of Learnt Hamiltonian Dynamics
  Inferred from Vision.
\newblock \emph{Advances in Neural Information Processing Systems}, 34.

\bibitem[{Holm, Schmah, and Stoica(2009)}]{holm2009geometric}
Holm, D.~D.; Schmah, T.; and Stoica, C. 2009.
\newblock \emph{Geometric mechanics and symmetry: from finite to infinite
  dimensions}, volume~12.
\newblock Oxford University Press.

\bibitem[{Kingma and Welling(2013)}]{kingma2013auto}
Kingma, D.~P.; and Welling, M. 2013.
\newblock Auto-encoding variational bayes.
\newblock \emph{arXiv preprint arXiv:1312.6114}.

\bibitem[{Kitagawa and Kakiuch(2017)}]{mocogan-pytorch}
Kitagawa, S.; and Kakiuch, K. 2017.
\newblock A pytorch implmention of {MoCoGAN}.
\newblock \url{https://github.com/DLHacks/mocogan}.

\bibitem[{Lutter, Ritter, and Peters(2019)}]{lutter2019deep}
Lutter, M.; Ritter, C.; and Peters, J. 2019.
\newblock Deep lagrangian networks: Using physics as model prior for deep
  learning.
\newblock \emph{arXiv preprint arXiv:1907.04490}.

\bibitem[{Rodas, Canal, and Taschin(2021)}]{rodas2021re}
Rodas, C.~B.; Canal, O.; and Taschin, F. 2021.
\newblock Re-Hamiltonian Generative Networks.
\newblock In \emph{ML Reproducibility Challenge 2020}.

\bibitem[{Roehrl et~al.(2020)Roehrl, Runkler, Brandtstetter, Tokic, and
  Obermayer}]{roehrl2020modeling}
Roehrl, M.~A.; Runkler, T.~A.; Brandtstetter, V.; Tokic, M.; and Obermayer, S.
  2020.
\newblock Modeling system dynamics with physics-informed neural networks based
  on Lagrangian mechanics.
\newblock \emph{IFAC-PapersOnLine}, 53(2): 9195--9200.

\bibitem[{Saemundsson et~al.(2020)Saemundsson, Terenin, Hofmann, and
  Deisenroth}]{saemundsson2020variational}
Saemundsson, S.; Terenin, A.; Hofmann, K.; and Deisenroth, M. 2020.
\newblock Variational integrator networks for physically structured embeddings.
\newblock In \emph{International Conference on Artificial Intelligence and
  Statistics}, 3078--3087. PMLR.

\bibitem[{Skorokhodov, Tulyakov, and Elhoseiny(2021)}]{skorokhodov2021stylegan}
Skorokhodov, I.; Tulyakov, S.; and Elhoseiny, M. 2021.
\newblock StyleGAN-V: A Continuous Video Generator with the Price, Image
  Quality and Perks of StyleGAN2.
\newblock \emph{arXiv preprint arXiv:2112.14683}.

\bibitem[{Tian et~al.(2021)Tian, Ren, Chai, Olszewski, Peng, Metaxas, and
  Tulyakov}]{tian2021good}
Tian, Y.; Ren, J.; Chai, M.; Olszewski, K.; Peng, X.; Metaxas, D.~N.; and
  Tulyakov, S. 2021.
\newblock A good image generator is what you need for high-resolution video
  synthesis.
\newblock \emph{arXiv preprint arXiv:2104.15069}.

\bibitem[{Toth et~al.(2019)Toth, Rezende, Jaegle, Racani{\`e}re, Botev, and
  Higgins}]{toth2019hamiltonian}
Toth, P.; Rezende, D.~J.; Jaegle, A.; Racani{\`e}re, S.; Botev, A.; and
  Higgins, I. 2019.
\newblock Hamiltonian generative networks.
\newblock \emph{arXiv preprint arXiv:1909.13789}.

\bibitem[{Tulyakov et~al.(2018)Tulyakov, Liu, Yang, and
  Kautz}]{tulyakov2018mocogan}
Tulyakov, S.; Liu, M.-Y.; Yang, X.; and Kautz, J. 2018.
\newblock Mocogan: Decomposing motion and content for video generation.
\newblock In \emph{Proceedings of the IEEE conference on computer vision and
  pattern recognition}, 1526--1535.

\bibitem[{Unterthiner et~al.(2019)Unterthiner, van Steenkiste, Kurach,
  Marinier, Michalski, and Gelly}]{unterthiner2019fvd}
Unterthiner, T.; van Steenkiste, S.; Kurach, K.; Marinier, R.; Michalski, M.;
  and Gelly, S. 2019.
\newblock FVD: A new metric for video generation.

\bibitem[{Van~der Maaten and Hinton(2008)}]{van2008visualizing}
Van~der Maaten, L.; and Hinton, G. 2008.
\newblock Visualizing data using t-SNE.
\newblock \emph{Journal of machine learning research}, 9(11).

\bibitem[{Vondrick, Pirsiavash, and Torralba(2016)}]{vondrick2016generating}
Vondrick, C.; Pirsiavash, H.; and Torralba, A. 2016.
\newblock Generating videos with scene dynamics.
\newblock \emph{Advances in neural information processing systems}, 29.

\bibitem[{Wang et~al.(2020{\natexlab{a}})Wang, Bilinski, Bremond, and
  Dantcheva}]{wang2020g3an}
Wang, Y.; Bilinski, P.; Bremond, F.; and Dantcheva, A. 2020{\natexlab{a}}.
\newblock G3AN: Disentangling appearance and motion for video generation.
\newblock In \emph{Proceedings of the IEEE/CVF Conference on Computer Vision
  and Pattern Recognition}, 5264--5273.

\bibitem[{Wang et~al.(2020{\natexlab{b}})Wang, Huang, Rudin, and
  Shaposhnik}]{wang2020understanding}
Wang, Y.; Huang, H.; Rudin, C.; and Shaposhnik, Y. 2020{\natexlab{b}}.
\newblock Understanding how dimension reduction tools work: an empirical
  approach to deciphering t-SNE, UMAP, TriMAP, and PaCMAP for data
  visualization.
\newblock \emph{arXiv preprint arXiv:2012.04456}.

\bibitem[{Wattenberg, Viégas, and Johnson(2016)}]{wattenberg2016how}
Wattenberg, M.; Viégas, F.; and Johnson, I. 2016.
\newblock How to Use t-SNE Effectively.
\newblock \emph{Distill}.

\bibitem[{Yoon, Jarrett, and Van~der Schaar(2019)}]{yoon2019time}
Yoon, J.; Jarrett, D.; and Van~der Schaar, M. 2019.
\newblock Time-series generative adversarial networks.
\newblock \emph{Advances in Neural Information Processing Systems}, 32.

\bibitem[{Yu et~al.(2022)Yu, Tack, Mo, Kim, Kim, Ha, and
  Shin}]{yu2022generating}
Yu, S.; Tack, J.; Mo, S.; Kim, H.; Kim, J.; Ha, J.-W.; and Shin, J. 2022.
\newblock Generating videos with dynamics-aware implicit generative adversarial
  networks.
\newblock \emph{arXiv preprint arXiv:2202.10571}.

\bibitem[{Zhong, Dey, and Chakraborty(2019)}]{zhong2019symplectic}
Zhong, Y.~D.; Dey, B.; and Chakraborty, A. 2019.
\newblock Symplectic ode-net: Learning hamiltonian dynamics with control.
\newblock \emph{arXiv preprint arXiv:1909.12077}.

\bibitem[{Zhong, Dey, and Chakraborty(2020)}]{zhong2020dissipative}
Zhong, Y.~D.; Dey, B.; and Chakraborty, A. 2020.
\newblock Dissipative symoden: Encoding hamiltonian dynamics with dissipation
  and control into deep learning.
\newblock \emph{arXiv preprint arXiv:2002.08860}.

\bibitem[{Zhong and Leonard(2020)}]{zhong2020unsupervised}
Zhong, Y.~D.; and Leonard, N. 2020.
\newblock Unsupervised learning of lagrangian dynamics from images for
  prediction and control.
\newblock \emph{Advances in Neural Information Processing Systems}, 33.

\end{thebibliography}
\end{document}